\documentclass[runningheads]{llncs}
\usepackage{graphicx}
\usepackage{comment}
\usepackage{amsmath,amssymb}
\usepackage{url,multirow,tabulary}
\usepackage{color}
\usepackage{subcaption}
\usepackage[cmintegrals]{newtxmath}
\usepackage{multirow}
\usepackage{makecell}
\usepackage{tabulary}
\usepackage{booktabs}
\usepackage{array}
\usepackage[table]{xcolor}
\usepackage{url}
\usepackage[width=122mm,left=12mm,paperwidth=146mm,height=193mm,top=12mm,paperheight=217mm]{geometry}
\newcommand{\ie}{\textit{i.e.}}

\newcommand{\etal}{\textit{et al.}}
\usepackage[colorlinks,bookmarks=false]{hyperref}
\usepackage{mathrsfs}
\usepackage{ifsym}
\usepackage{bbding}

\newcolumntype{L}[1]{>{\raggedright\arraybackslash}p{#1}}
\newcolumntype{C}[1]{>{\centering\arraybackslash}p{#1}}
\newcolumntype{R}[1]{>{\raggedleft\arraybackslash}p{#1}}

\usepackage{pifont}
\newcommand{\cmark}{\ding{51}}

\begin{document}
\pagestyle{headings}\mainmatter\def\ECCVSubNumber{1}

\title{ATG-PVD: Ticketing Parking\\ Violations on A Drone}

\titlerunning{ATG-PVD: Ticketing Parking Violations on A Drone}
\authorrunning{H. Wang \etal}

\author{Hengli Wang\inst{1}$^\star$ \and
    Yuxuan Liu\inst{1}$^\star$ \and Huaiyang Huang\inst{1}$^\star$ \and Yuheng Pan\inst{2}\thanks{Equal contributions.} \and\\Wenbin Yu\inst{2} \and Jialin Jiang\inst{2} \and Dianbin Lyu\inst{2} \and Mohammud J. Bocus\inst{3}\and\\Ming Liu\inst{1} \and Ioannis Pitas\inst{4} \and Rui Fan\inst{2,5}\Envelope}
    \institute{HKUST Robotics Institute\\
    \email{\{hwangdf,yliuhb,hhuangat,eelium\}@ust.hk}
    \\
    \and
    ATG Robotics\\
    \email{\{panyuheng,yuwenbin,jiangjialin,lvdianbin\}@atg-itech.com}\\
    \and
    University of Bristol\\
    \email{junaid.bocus@bristol.ac.uk}\\
    \and
    Aristotle University of Thessaloniki\\
    \email{pitas@csd.auth.gr}\\
    \and UC San Diego\\
    \email{rui.fan@ieee.org}\\
}

\maketitle

\begin{abstract}
In this paper, we introduce a novel suspect-and-investigate framework, which can be easily embedded in a drone for automated parking violation detection (PVD). Our proposed framework consists of: 1) SwiftFlow, an efficient and accurate convolutional neural network (CNN) for unsupervised optical flow estimation; 2) Flow-RCNN, a flow-guided CNN for car detection and classification; and 3) an illegally parked car (IPC) candidate investigation module developed based on visual SLAM. The proposed framework was successfully embedded in a drone from ATG Robotics. The experimental results demonstrate that, firstly, our proposed SwiftFlow outperforms all other state-of-the-art unsupervised optical flow estimation approaches in terms of both speed and accuracy; secondly, IPC candidates can be effectively and efficiently detected by our proposed Flow-RCNN, with a better performance than our baseline network, Faster-RCNN; finally, the actual IPCs can be successfully verified by our investigation module after drone re-localization.
\end{abstract}

\begin{center}
    \textbf{Dataset and Demo Video:}\\
    \url{sites.google.com/view/atg-pvd}
\end{center}

\section{Introduction}
\label{sec.introduction}
We are currently experiencing an unprecedented crisis due to the ongoing Coronavirus Disease 2019 (COVID-19) pandemic. Its worldwide escalation has taken us by surprise, causing major disruptions to global health, economic and social systems. Indeed, our lives have changed overnight -- businesses and schools are closed, most employees are working from home, and many have found themselves without a job. Millions of people across the globe are confined to their homes, while healthcare workers are at the frontline of the COVID-19 response \cite{mckee2020if}. With the increase in COVID-19 cases, public transport use has plummeted, as commuters shun buses, trams, and trains in favor of private cars and taxis. For instance, USA Today reported that the transit ridership demand in April 2020 was down by about 75\% nationwide, compared to normal, with figures of 85\% in San Francisco, 67\% in Detroit and 60\% in Philadelphia \cite{hughes2020usatoday}.

With the increasing number of vehicles on the roads, parking spaces have become scarce and many vehicles are parked just by the roadside, which in turn results in a significant increase in parking violations. In late March 2020, the Department of Transportation in Los Angeles \cite{relaxed_parking2020} announced relaxed parking enforcement regulations as part of the emergency response to COVID-19, so that their citizens could practice safe social distancing without being concerned about a ticket. As the Return-to-Work Plan progresses, the relaxed parking enforcement regulations are no longer in force, consequently increasing the workload of the local traffic law enforcement officers. The demand for automated and intelligent parking violation detection (PVD) systems has thus become greater than ever.

\begin{figure*}[!t]
    \centering
    \includegraphics[width=0.99\textwidth]{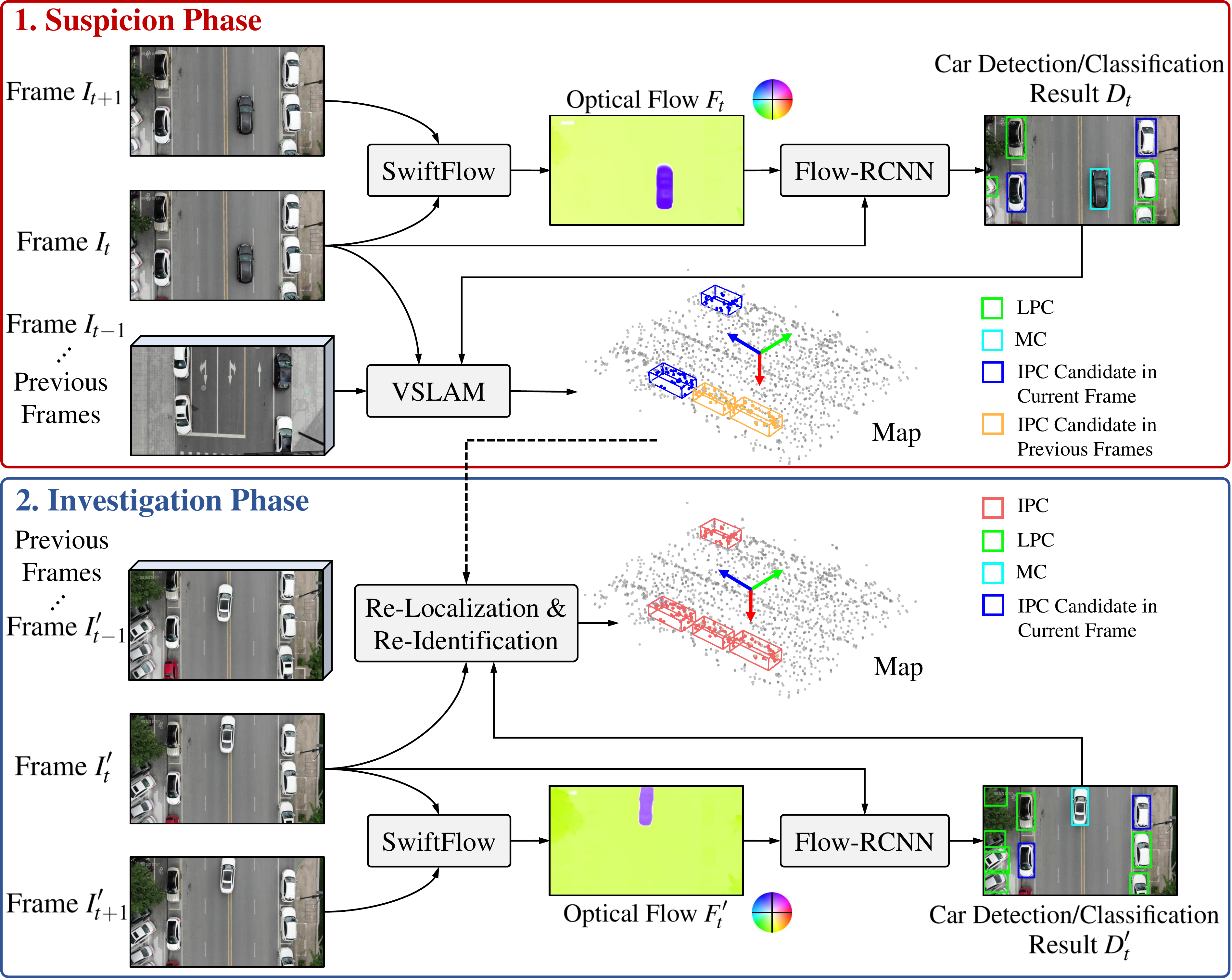}
    \caption{The framework of our proposed suspect-and-investigate PVD system: the first phase identifies suspected IPC candidates, and the second phase investigates the suspected IPC candidates and issues tickets to the actual IPCs. The frame $I_{t}$ in the suspicion phase corresponds to the frame $I_{t}'$ in the investigation phase.}
    \label{fig.framework}
\end{figure*}

The existing automated PVD systems typically recognize illegally parked cars (IPCs) by analyzing the videos acquired by closed-circuit televisions (CCTVs) through 2D/3D object detection algorithms \cite{zhou2016dave} or video surveillance analysis algorithms \cite{regazzoni2010video}. However, the efficiency of such methods relies on CCTV camera positions, as IPCs cannot always be detected, especially if they are at a distant location. Deploying more CCTVs can definitely minimize misdetections, but this will also incur a high cost, and/or may not be practical. Therefore, many researchers have turned their focus towards mobile PVD systems, which can be mounted on any vehicle type. For example, the Birmingham City Council in England utilizes surveillance cars to detect IPCs and record their plate numbers \cite{council2013codes}. However, such surveillance cars are expensive and typically require drivers. Therefore, autonomous machines, especially drones, have emerged as more efficient and cheaper alternatives.

The cars in the street can be grouped into three categories: 1) moving cars (MCs), 2) legally parked cars (LPCs) and 3) IPCs. MCs can be distinguished from LPCs and IPCs using dynamic object detection techniques, such as optical flow analysis, while IPCs can be distinguished from LPCs using object detection networks, such as Faster-RCNN \cite{FasterRCNN}, with the assistance of parking spot  information. In this paper, we introduce a novel \textit{suspect-and-investigate} PVD system (see Fig. \ref{fig.framework}) embedded in a drone. In the suspicion phase, we first employ a novel unsupervised optical flow estimation network, referred to as \textit{SwiftFlow}, to estimate the optical flow $F_{t}$ between $I_{t}$ and $I_{t+1}$. $F_{t}$ is then incorporated into a novel object detection and classification network, referred to as \textit{Flow-RCNN}, to detect cars and classify them into MCs, LPCs and IPC candidates. A visual simultaneous localization and mapping (VSLAM) module then builds a localizable map containing the suspected IPC candidates. After a parking grace period (which is typically five minutes) has elapsed, the drone flies back to the same location. The VSLAM module in the investigation phase subsequently detects loop closure and re-localizes the drone in the pre-built map. Finally, the suspected IPC candidates are re-identified, and the actual IPCs are marked in the map. Our main contributions are summarized as follows:
\begin{itemize}
    \item A novel suspect-and-investigate PVD framework;
    \item SwiftFlow, a novel unsupervised optical flow estimation network;
    \item Flow-RCNN, a novel car detection and classification network;
    \item A large-scale PVD dataset, published for research purposes.
\end{itemize}

\section{Related Work}
\label{sec.related_work}

\subsection{Optical Flow Estimation}
\label{sec.optical_flow_estimation}
Traditional approaches generally formulate optical flow estimation as a global energy minimization problem \cite{horn1981determining,memin1998dense,brox2004high,zach2007duality}. Recently, convolutional neural networks (CNNs) have achieved impressive performance in optical flow estimation. FlowNet \cite{dosovitskiy2015flownet} was the pioneering work in end-to-end deep optical flow estimation. Its key component is a so-called correlation layer, which can provide explicit matching capabilities. Later methods, PWC-Net \cite{sun2018pwc} and LiteFlowNet \cite{hui2018liteflownet} introduced the popular coarse-to-fine architecture, which provides a good trade-off between optical flow accuracy and computation efficiency. Meanwhile, IRR-PWCNet \cite{hur2019iterative} demonstrates that occlusion prediction integrated into optical flow estimation can effectively enhance the optical flow estimation accuracy.

Although the aforementioned supervised optical flow estimation methods perform impressively, they generally require a large amount of optical flow ground truth to learn the best solution. Acquiring such ground truth, especially for real-world datasets, is extremely time-consuming and labor-intensive, making these supervised approaches difficult to apply in real-world applications. For these reasons, unsupervised learning has recently become the preferred technique for such applications. For instance, DSTFlow \cite{ren2017unsupervised} employs a photometric loss and a smooth loss in CNN training, which are similar to the global energy used in traditional methods. Additionally, some methods, such as UnFlow \cite{meister2018unflow}, DDFlow \cite{liu2019ddflow} and SelFlow \cite{liu2019selflow} integrate occlusion reasoning into unsupervised optical flow estimation frameworks to further improve their accuracy. However, such approaches are typically computationally intensive, and they are difficult to embed in a drone.

\subsection{Object Detection}
\label{sec.object_detection}
Discovering objects and their locations in images is still a challenging problem in  computer vision. Due to their promising results, CNNs have emerged as a powerful tool for object detection. The modern deep object detection algorithms can be grouped into two main types: a) anchor-based and b) anchor-free.

Anchor-based methods predict bounding boxes based on initial guesses. According to the pipelines and primary proposal sources, they can be further categorized as either one-stage or two-stage methods. The former make predictions directly from hand-crafted anchors. For example, RetinaNet \cite{focalLoss} employs a feature pyramid network (FPN) to produce dense predictions at multiple scales.
On the other hand, the two-stage methods make predictions using the proposals produced by a one-stage detector. For instance, Fast-RCNN \cite{girshickICCV15fastrcnn} and Faster-RCNN \cite{FasterRCNN} perform cropping and resizing on images or feature maps, according to the bounding box proposals. The RCNN branch in Faster-RCNN utilizes a field of view (FOV), that is larger than the bounding box proposals, so as to extract regions of interest (RoIs) directly from the feature maps.

Anchor-free methods usually do not rely on human-designed region proposals to bootstrap the detection process. For example, CornerNet \cite{Law_2018_CornerNet} translates the object detection problem into a keypoint detection and matching problem, where specially-designed pooling layers construct biased receptive fields for corner point detection. CenterNet \cite{zhou2019objects}, which is based on CornerNet \cite{Law_2018_CornerNet},  utilizes two customized modules: a) cascade corner pooling and b) center pooling, to enrich information collected by both the top-left and bottom-right corners. It detects each object as a triplet, rather than a pair, of keypoints.

In recent years, incorporating additional visual information, such as semantic predictions, into object classification is becoming  an increasingly ubiquitous part of object detection. Since MCs can be easily distinguished from optical flow images, we incorporate the latter into our framework to improve IPC candidate detection.

\subsection{VSLAM}
\label{sec.slam}
Traditional VSLAM approaches leverage visual features and the geometric relations between multiple views of a 3D scene (typically known as multi-view geometry) to estimate camera poses and construct/update a map of the 3D scene. The state-of-the-art VSLAM approaches are classified as either indirect \cite{klein2007ptam,strasdat2011double,mur2015orb} or direct \cite{newcombe2011dtam,engel2014lsd,engel2018dso}. Both types extract visual features from images and associate them with descriptors. However, the indirect methods sample corners and associate them with higher dimensional descriptors, while the direct methods typically sample pixels with a relatively large local intensity gradient and associate them with a patch of pixels surrounding their sampled location. Furthermore, these two types of methods typically minimize different objective functions: the indirect methods resort to geometric residuals, whereas the direct methods resort to photometric residuals.

In order to combine the advantages of these two types of methods, Froster \textit{et al}. \cite{forster2014svo} proposed semi-direct visual odometry (SVO), which tracks camera poses via sparse image alignment and utilizes hierarchical bundle adjustment (BA) as the back-end to optimize the geometry structure and camera motion. Furthermore, many researchers have integrated other computer vision tasks, such as 2D object detection \cite{yang2019cubeslam,huang2020clustervo,nicholson2018quadricslam}, instance segmentation \cite{runz2018maskfusion,mccormac2018fusion++} and flow/depth prediction \cite{zhang2020flowfusion,tateno2017cnn},  into their SLAM systems, so as to address the problem of the existence of dynamic objects, by exploiting high-level semantic information. For example, Huang \textit{et al}. \cite{huang2020clustervo} proposed ClusterVO, which uses a multi-level probabilistic association scheme to both track low-level visual features and realize high-level object detection. Moreover, Yang \textit{et al}. \cite{yang2019cubeslam} introduced CubeSLAM, which performs single image 3D cuboid object detection, together with multi-view object SLAM.

\section{ATG-PVD Framework}
\label{sec.methodology}

\begin{figure*}[!t]
    \centering
    \includegraphics[width=0.99\textwidth]{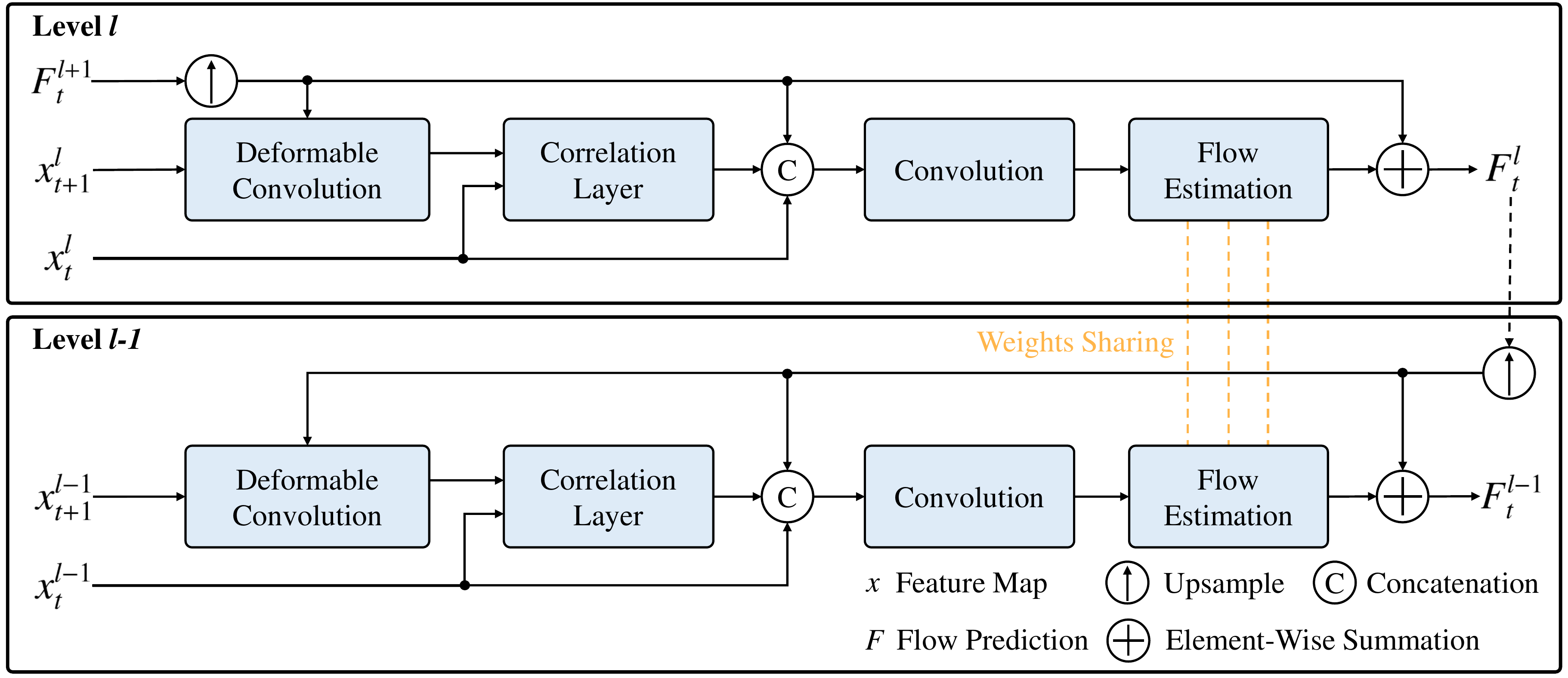}
    \caption{The decoder architecture of our proposed SwiftFlow. The pipeline of two adjacent levels in the decoder are displayed for simplicity.}
    \label{fig.swiftflow}
\end{figure*}

\subsection{SwiftFlow}
\label{sec.unsupervised_optical_flow_estimation}
Since our proposed SwiftFlow network is based on the pipeline of PWC-Net \cite{sun2018pwc}, we first provide readers with some preliminaries about the latter. In PWC-Net \cite{sun2018pwc}, feature maps are first extracted from video frames using a Siamese pyramid network. Then, the feature map $x_{t+1}^{l}$ of the ${(t+1)}$-th video frame at level $l$ is aligned with the feature map $x_{t}^{l}$ of the ${t}$-th video frame at level $l$ via a warping operation based on the upsampled flow prediction $F_{t}^{l+1}$ at level $l+1$. A correlation layer is then employed to compute the cost volume, which is subsequently concatenated with $x_{t}^{l}$ as well as the upsampled flow prediction $F_{t}^{l+1}$ at level $l+1$. Finally, the flow residual, predicted by the flow estimation module, is combined with the upsampled flow prediction $F_{t}^{l+1}$ at level $l+1$ using an element-wise summation to generate the flow prediction $F_{t}^{l}$ at level $l$. We iterate this process and obtain the flow predictions at different scales.

SwiftFlow improves on PWC-Net \cite{sun2018pwc} in terms of computational efficiency, so that it can perform in real time on a drone. The decoder in PWC-Net \cite{sun2018pwc} has too many parameters, so we make three major modifications to the decoder architecture (see Fig. \ref{fig.swiftflow}) to minimize the model size and improve accuracy. As the decoder in PWC-Net \cite{sun2018pwc} employs a dense connection scheme in each pyramid level, making the network computationally intensive, SwiftFlow establishes connections only between two adjacent levels, which can reduce the number of network parameters by 50\%. Furthermore, the optical flow estimation modules at different pyramid levels of PWC-Net \cite{sun2018pwc} have different learnable weights to estimate optical flow residuals. Considering that the optical flow estimation modules at different levels have the same functionality and the optical flow residuals at different levels have similar value ranges, we believe sharing the weights of optical flow estimation modules at all pyramid levels can be a more effective and efficient strategy. We also add an additional convolutional layer before the optical flow estimation module at each level for feature map alignment. Moreover, we notice that the warping operation can induce ambiguity to occluded areas, which breaks correlation layer symmetry. We propose to add an asymmetric layer before the correlation layer to alleviate this problem and improve optical flow estimation accuracy. Therefore, we replace the warping operation with a deformable convolutional layer \cite{dai2017deformable}, as shown in Fig.~\ref{fig.swiftflow}.

Referring to the commonly applied unsupervised training strategy, we train SwiftFlow by minimizing the following weighted sum of losses:
\begin{equation}
    L=\lambda_{\text{photo}} \cdot L_{\text{photo}}+\lambda_{\text{smooth}} \cdot L_{\text{smooth}}+\lambda_{\text{self}} \cdot L_{\text{self}},
\end{equation}
where $L_{\text{photo}}$ is the photometric loss that considers an occlusion-aware mask \cite{wang2018occlusion}, $L_{\text{smooth}}$ is the smoothness regularization \cite{tomasi1998bilateral}, and $L_{\text{self}}$ is the self-supervision Charbonnier loss \cite{liu2019ddflow}. Following the instructions in \cite{jonschkowski2020matters}, we set $\lambda_{\text{photo}}=1$ and $\lambda_{\text{smooth}}=2$ in our experiments. Moreover, we use $\lambda_{\text{self}}=0$ for the first $50\%$ of training steps, and increase it to 0.3 linearly for the next $10\%$ of training steps, after which it stays at a constant value.

\subsection{Flow-RCNN}
\label{sec.parking_violator_candidate_detection}
Given an RGB video frame and its corresponding estimated optical flow, the proposed Flow-RCNN detects cars in the video frame and classifies them into MCs, LPCs, and IPC candidates.

\begin{figure}[!t]
    \centering
    \includegraphics[width=0.990\columnwidth]{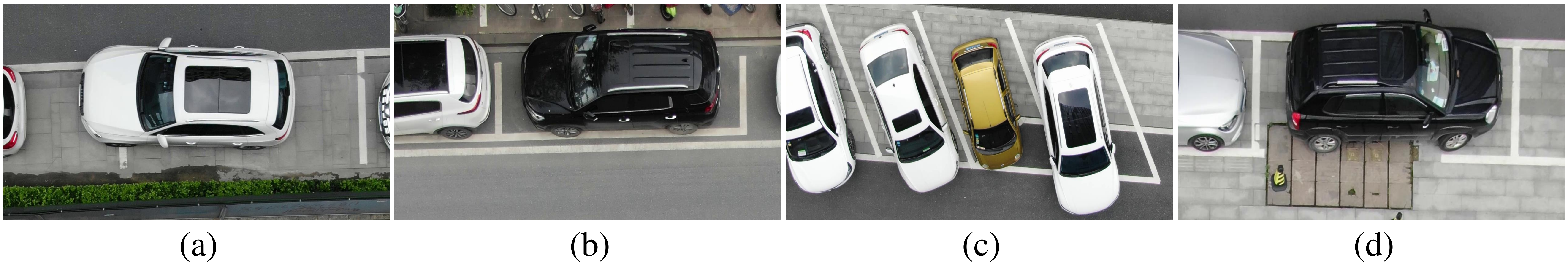}
    \caption{Challenging cases for parked car detection and classification.}
    \label{fig.CommonScenes}
\end{figure}

Judging whether a car is legally parked is very challenging. Intuitively, we can resort to the parking spot delimitation lines, which are typically painted in white. However, in real-world environments,  methods that rely solely on the parking spot information may fail. For instance, in Fig. \ref{fig.CommonScenes}(a), the white car is not parked entirely within the designated parking spot; in Fig. \ref{fig.CommonScenes}(b), only parts of the white car and parking spot appear; and in Fig. \ref{fig.CommonScenes}(c) and Fig. \ref{fig.CommonScenes}(d), the parking spots are not enclosed. Moreover, parking spots are not always bounded by rectangular line markings, as illustrated in Fig. \ref{fig.CommonScenes}(c). It is challenging to design a rule-guided method to solve for these cases, even with perfectly labeled cars and parking spots. Furthermore, various tall objects, such as light poles and trees, often present salient optical flow estimations. In this case, the methods that rely entirely on optical flow information can wrongly characterize an IPC/LPC as an MC. Therefore, an end-to-end, optical flow-guided, and detect-and-classify architecture for IPC candidate detection provides a better alternative.

\begin{figure}[!t]
    \centering
    \includegraphics[width=0.99\columnwidth]{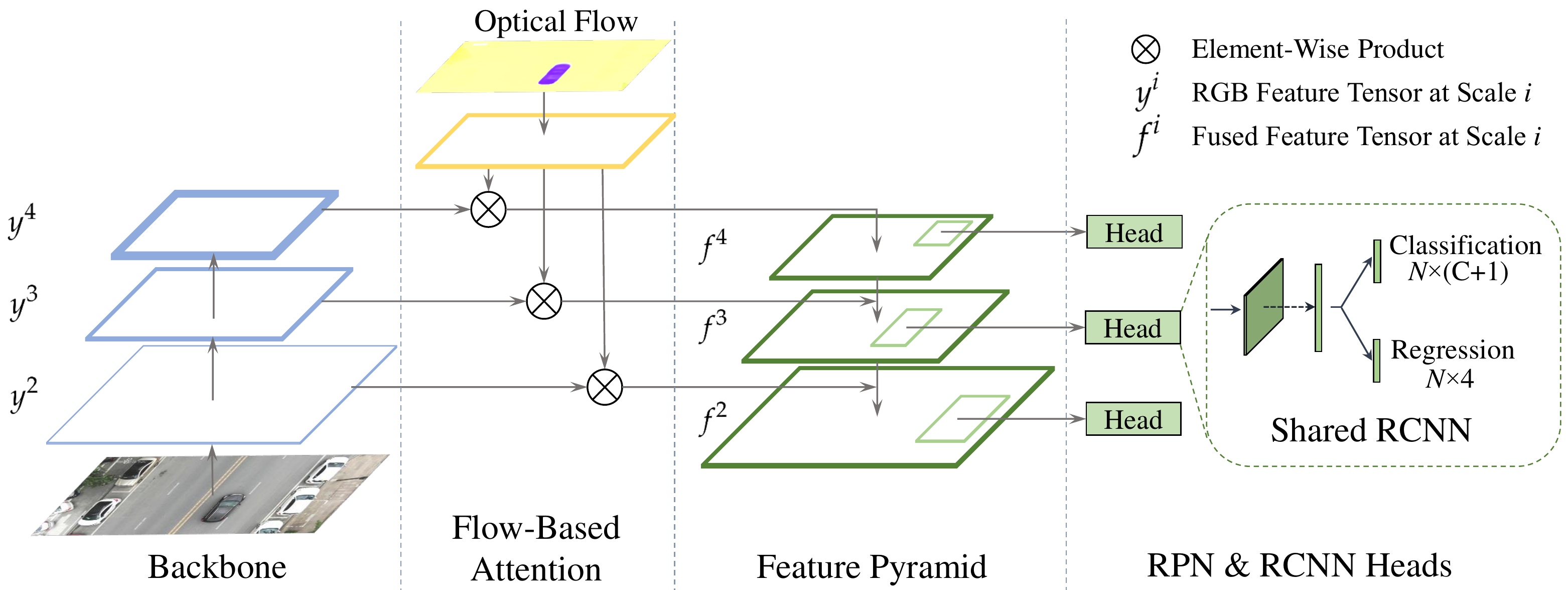}
    \caption{Flow-RCNN architecture. The optical flow image, obtained from SwiftFlow, is fed into multiple convolutional layers. The optical flow features then dynamically weigh each element in the multi-scale feature maps extracted from the RGB image.}
    \label{fig.FasterRCNN_arch}
\end{figure}

The architecture of our proposed Flow-RCNN is illustrated in Fig. \ref{fig.FasterRCNN_arch}. It incorporates the optical flow information, obtained by SwiftFlow in Section \ref{sec.unsupervised_optical_flow_estimation}, into the conventional Faster-RCNN \cite{FasterRCNN} architecture for IPC candidate detection, and it outputs the position and category (MC, LPC or IPC candidate) of each car in the video frame in an end-to-end manner. The RGB video frame is first passed through a backbone CNN to produce multi-scale feature maps $y^i$. The features extracted from the optical flow image then dynamically weigh the activation of each element in the multi-scale feature maps $y^i$, which enables the detector to focus more on MCs.
We then fuse the multi-scale feature maps to produce a feature pyramid for the subsequent region proposal network (RPN) and RCNN heads \cite{FasterRCNN}.
Since our dataset is highly imbalanced (see Fig. \ref{fig.data_example}), \ie, most vehicles are regarded as IPC candidates or IPCs, we apply focal loss \cite{focalLoss} to mitigate the class imbalance problem in the classification stage.

\subsection{Mapping, Re-Localization and Re-Identification}
\label{sec.mapping_relocalization_and_reidentification}
\begin{figure}[!t]
    \centering
    \includegraphics[width=0.99\textwidth]{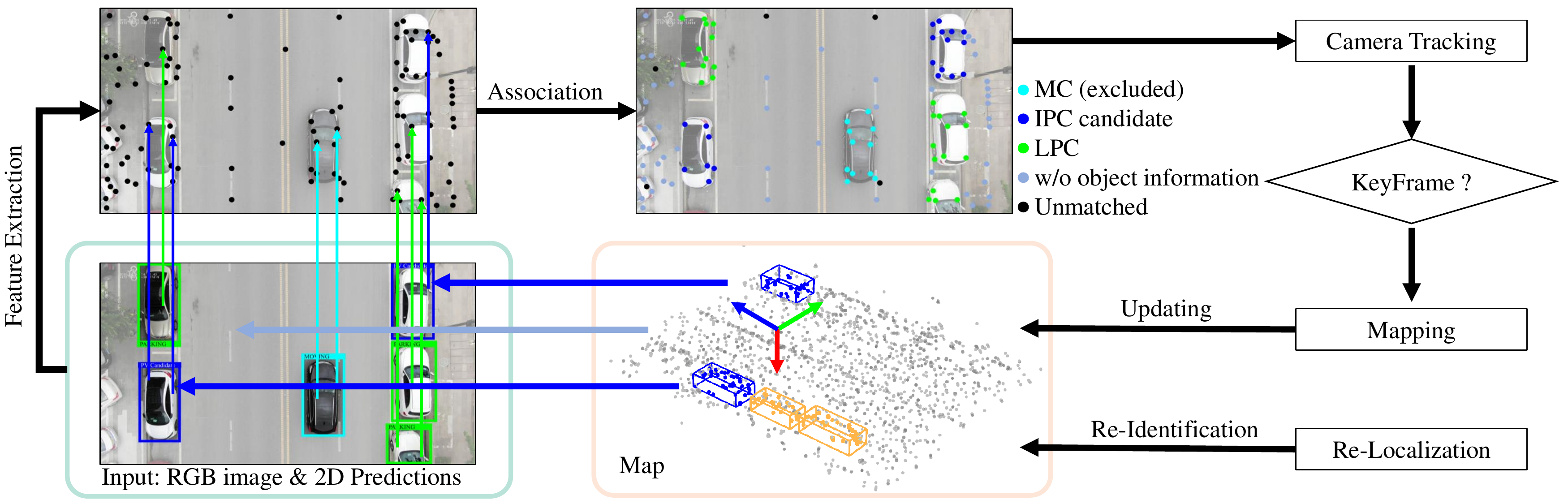}
    \caption{The pipeline of our mapping, re-localization and re-identification module.}
    \label{fig.mapping_pipeline}
\end{figure}

Given RGB images and the corresponding detected IPC candidates, our next target is to build a 3D map, investigate each IPC candidate and mark it in the map. To this end, we develop a mapping, re-localization and re-identification module, as illustrated in Fig.~\ref{fig.mapping_pipeline}, on top of ORB-SLAM2 \cite{mur2017orb}.

Our proposed system applies a suspect-and-investigate scheme to mark IPCs in 3D.
In the suspicion phase, we leverage ORB-SLAM2 \cite{mur2017orb} to build a 3D localizable map and mark the detected IPC candidates in the map.
Given an RGB image containing detected IPC candidates,
the system first extracts ORB \cite{rublee2011orb} features $\left\{\mathbf{u}_0, \dots,  \mathbf{u}_t \right\}$ and associates them with 2D bounding boxes $\left\{\mathcal{B}^{2D}_0, \dots, \mathcal{B}^{2D}_h \right\}$.
We explicitly exclude the ORB features extracted from MCs in the subsequent procedures, \ie, tracking and mapping.
The rest of the features are then matched with the 3D keypoints $\left\{\mathbf{x}_0, \dots, \mathbf{x}_m\right\}$ in the map.
With these 3D-2D correspondences $\mathcal{K} \doteq \left\{(i_k, j_k)\right\}_{k=1:N}$, the current camera pose $\mathbf{T} = [\mathbf{R}, \mathbf{t}]$ is estimated in a perspective-n-point (PnP) scheme by minimizing the reprojection error as follows \cite{mur2017orb}:
\begin{equation}
    \mathbf{R^\ast}, \mathbf{t^\ast} = \arg \min_{\mathbf{R}, \mathbf{t}} \sum_{(i, j) \in \mathcal{K}}\| \mathbf{u}_i - \pi(\mathbf{R}\mathbf{x}_j - \mathbf{t})\|,
\end{equation}
where $\pi (\cdot)$ is the camera projection function.
After solving the camera pose, the inlier correspondences $\mathcal{K}^\ast \doteq \left\{(i_k, j_k)\right\}_{k=1:N^\prime}$
can be determined via their reprojection errors.
Then we attempt to associate 2D bounding boxes in the current frame with candidates in the map.
A pair of 3D and 2D bounding boxes $(\mathcal{B}^{3D}_i, \mathcal{B}^{2D}_j )$ is associated if $|\mathcal{K}_{ij}| > \delta_\text{obj}$, where $\mathcal{K}_{ij}$ is a subset of $\mathcal{K}$, $(\mathbf{u}_i, \mathbf{x}_j)$ with $(i, j) \in \mathcal{K}^\ast$ is a pair of 2D/3D keypoints belonging to a pair of 2D/3D bounding boxes respectively and $\delta_\text{obj}$ is the threshold.
In the mapping module, the system triangulates 2D feature correspondences into 3D keypoints, which are assigned with their corresponding 3D bounding box information.
Then, it jointly optimizes the camera poses of keyframes $\left\{\mathbf{T}_0, \dots, \mathbf{T}_n\right\}$ and the 3D keypoint positions $\left\{\mathbf{x}_0, \dots, \mathbf{x}_m\right\}$.
We consider the 3D bounding boxes in the suspicion phase as IPC candidates and mark them in the map.
In the investigation phase, the system detects loop closure to re-localize the drone in the pre-built map.
After the drone is successfully re-localized, we further verify existing IPC candidates.
In the re-localization stage, if sufficient semantic keypoints belonging to a candidate $\mathcal{B}^{3D}_i$ are associated with a detected vehicle $\mathcal{B}^{2D}_j$ in the current frame, we re-identify the candidate as an IPC and mark it in the map. The proposed solution does not take into account that the local traffic law enforcement officers already have 2D street maps with labeled parking spots, but the drone map can be registered with such 2D street maps to greatly improve IPC detection.

\section{Experiments}
\label{sec.experimental_results}

\subsection{Experimental Setup}
\label{sec.exp_setup}

\begin{figure}[!t]
    \centering
    \includegraphics[width=\textwidth]{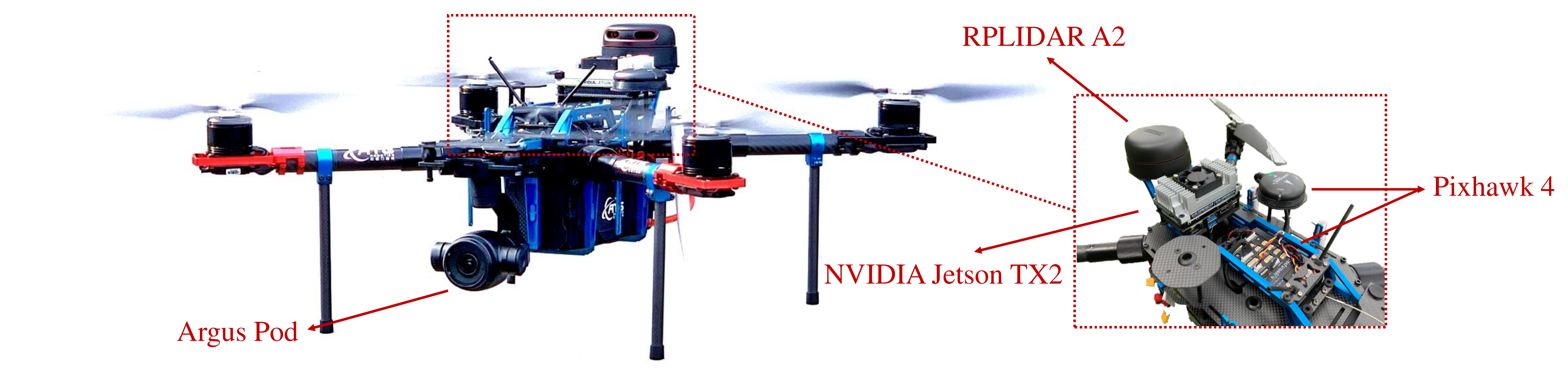}
    \caption{Experimental setup.}
    \label{fig.uav_setup}
\end{figure}

Our proposed PVD system is embedded in an ATG-R680 drone\footnote{\url{atg-itech.com}} (see Fig. \ref{fig.uav_setup}), controlled by a Pixhawk 4\footnote{\url{docs.px4.io/v1.9.0/en/flight_controller/pixhawk4.html}} advanced autopilot. The maximum take-off weight of the drone is 5.6 kg. We utilize an Argus zoom pot\footnote{\url{topxgun.com/en/product-argus.html}} microminiature tri-axis gimbal camera to capture images with a resolution of $2160\times3840$ pixels at 25 fps. The captured images are then processed by an NVIDIA Jetson TX2 GPU\footnote{\url{developer.nvidia.com/embedded/jetson-tx2}}, which has an 8 GB LPDDR4 memory and 256 CUDA cores, for IPC detection. Furthermore, we also equip our drone with an RPLIDAR A2\footnote{\url{slamtec.com/en/Lidar/A2}}, which can perform 360$^\circ$ omnidirectional laser range scanning.

\subsection{ATG-PVD Dataset}
\label{sec.dataset}

\begin{figure}[!t]
    \centering
    \includegraphics[width=1.0\linewidth]{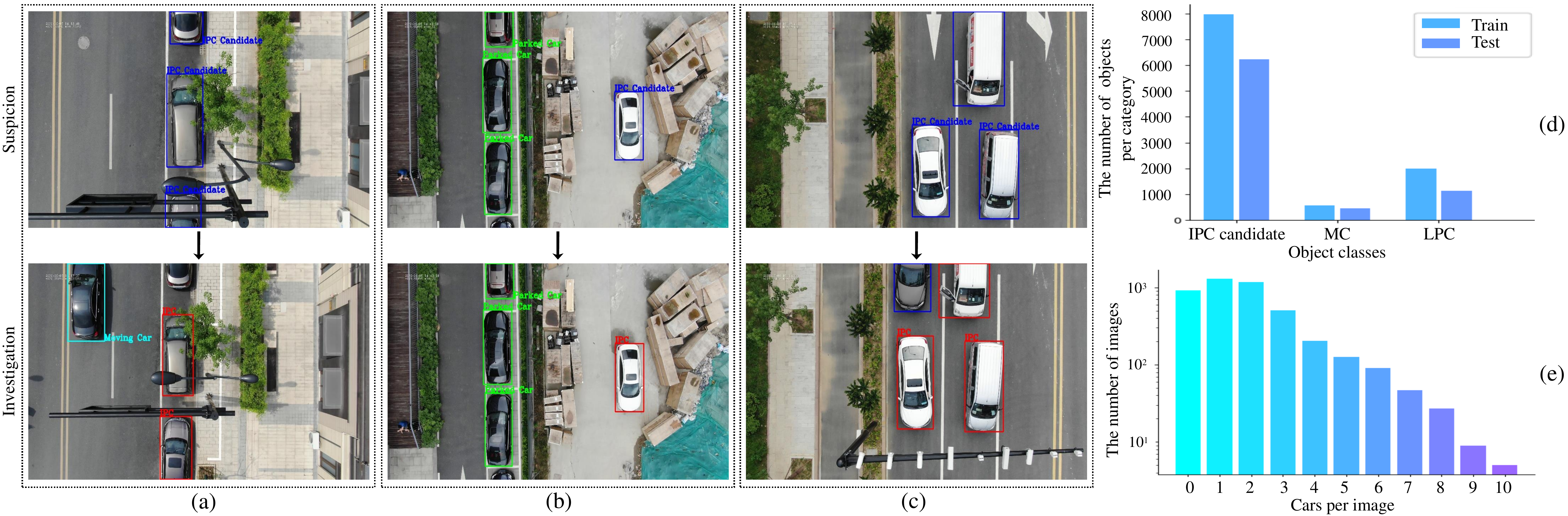}
    \caption{Our created ATG-PVD dataset: (a)--(c) the images on the first row are used in the suspicion phase, while the images on the second row are used in the investigation phase; (d) and (e) the statistical analysis of the dataset.}
    \label{fig.data_example}
\end{figure}

Using the aforementioned experimental setup, we created a large-scale real-world dataset, named the ATG-PVD dataset, for parking violation detection. Our dataset is publicly available at \url{sites.google.com/view/atg-pvd} for research purposes.
The ATG-PVD dataset contains seven sequences (resolution: $2160 \times 3840$ pixels) and the corresponding 2D bounding box annotations for car detection and classification. The ground truth used in the suspicion phase has three classes: a) IPC candidates, b) MCs and c) LPCs, while in the investigation phase, the IPC ground truth is also provided. Examples of the images used in the suspicion and investigation phases are shown in Fig.~\ref{fig.data_example}(a)--(c).

In our experiments, we divide our ATG-PVD dataset into a training set and a testing set, which respectively contains 4924 and 4398 images. The statistics for these two sets are shown in Fig.~\ref{fig.data_example}(d) and (e), where it can be observed that there are more IPC candidates or IPCs than MCs and LPCs. Additionally, most images contain fewer than five cars. Furthermore, our experiments are conducted on downsampled images with a resolution of $540 \times 960$ pixels.
Sections \ref{sec.evaluations_of_optical_flow_estimation}, \ref{sec.evaluations_of_parking_violator_candidate_detection}, and \ref{sec.evaluations_of_the_parking_violator_detection_system} respectively discuss the performances of SwiftFlow, Flow-RCNN and our PVD system in terms of both qualitative and quantitative experimental results.

\begin{table}[!t]
    \caption{Ablation study of our SwiftFlow on the KITTI flow 2015 \cite{kitti15} training dataset. Best results are shown in bold font.}
    \centering
    \begin{tabular}{L{1.5cm}C{1.7cm}C{1.7cm}C{2.2cm}C{1.4cm}C{1.7cm}}
        \toprule
        \multicolumn{1}{c}{Backbone} & Reduce Dense & Shared Weights & Deformable Convolution & F1-all ($\%$) & \# Params (M) \\ \midrule
        \multirow{4}{*}{PWC-Net} & -- & -- & -- & 8.37 & 8.75 \\
        & \cmark & -- & -- & 7.22 & 5.26 \\
        & \cmark & \cmark & -- & 6.95 & \textbf{2.18} \\
        & \cmark & \cmark & \cmark & \textbf{6.51} & 2.51 \\
        \bottomrule
    \end{tabular}
    \label{tab.swiftflow_ablation}
\end{table}

\begin{table*}[!t]
    \caption{The evaluation results on the KITTI flow benchmarks, where DDFlow \cite{liu2019ddflow}, UnFlow \cite{meister2018unflow}, Flow2Stereo \cite{liu2020flow2stereo} and SelFlow \cite{liu2019selflow} are the state-of-the-art self-supervised approaches. Best results are shown in bold font.}
    \centering
    \begin{tabular}{L{3cm}C{2cm}C{1.25cm}C{2cm}C{1.25cm}C{2cm}}
        \toprule
        \multicolumn{1}{l}{\multirow{2}{*}{Approach}} & \multicolumn{2}{c}{KITTI 2012} & \multicolumn{2}{c}{KITTI 2015} & \multicolumn{1}{c}{\multirow{2}{*}{Runtime (s)}} \\ \cmidrule(l){2-3} \cmidrule(l){4-5}
        \multicolumn{1}{c}{} & Out-Noc~($\%$) & Rank & F1-all~($\%$) & Rank & \multicolumn{1}{c}{} \\ \midrule
        DDFlow \cite{liu2019ddflow} & 4.57 & 60 & 14.29 & 91 & 0.06 \\
        UnFlow \cite{meister2018unflow} & 4.28 & 53 & 11.11 & 66 & 0.12 \\
        Flow2Stereo \cite{liu2020flow2stereo} & 4.02 & 48 & 11.10 & 65 & 0.05 \\
        SelFlow \cite{liu2019selflow} & 3.32 & 34 & 8.42 & 51 & 0.09 \\ \midrule
        SwiftFlow (Ours) & \textbf{2.64} & \textbf{24} & \textbf{7.23} & \textbf{35} & \textbf{0.03} \\ \bottomrule
    \end{tabular}
    \label{tab.kitti}
\end{table*}

\subsection{Evaluation of SwiftFlow}
\label{sec.evaluations_of_optical_flow_estimation}

\subsubsection{Ablation Study}
\label{sec.ablation_study}

We conduct an ablation study to validate the effectiveness of SwiftFLow. The experimental results are presented in Table \ref{tab.swiftflow_ablation}. We can see that, by removing dense connections between different levels, our approach can reduce many parameters, but still retain a similar optical flow estimation performance, compared with the PWC-Net \cite{sun2018pwc} baseline. Moreover, sharing weights of flow estimation modules can yield a performance improvement with fewer parameters. Furthermore, thanks to deformable convolution, our proposed SwiftFlow achieves the best performance with only a few additional parameters.
\subsubsection{Evaluation}

\begin{figure}[!t]
    \centering
    \includegraphics[width=\textwidth]{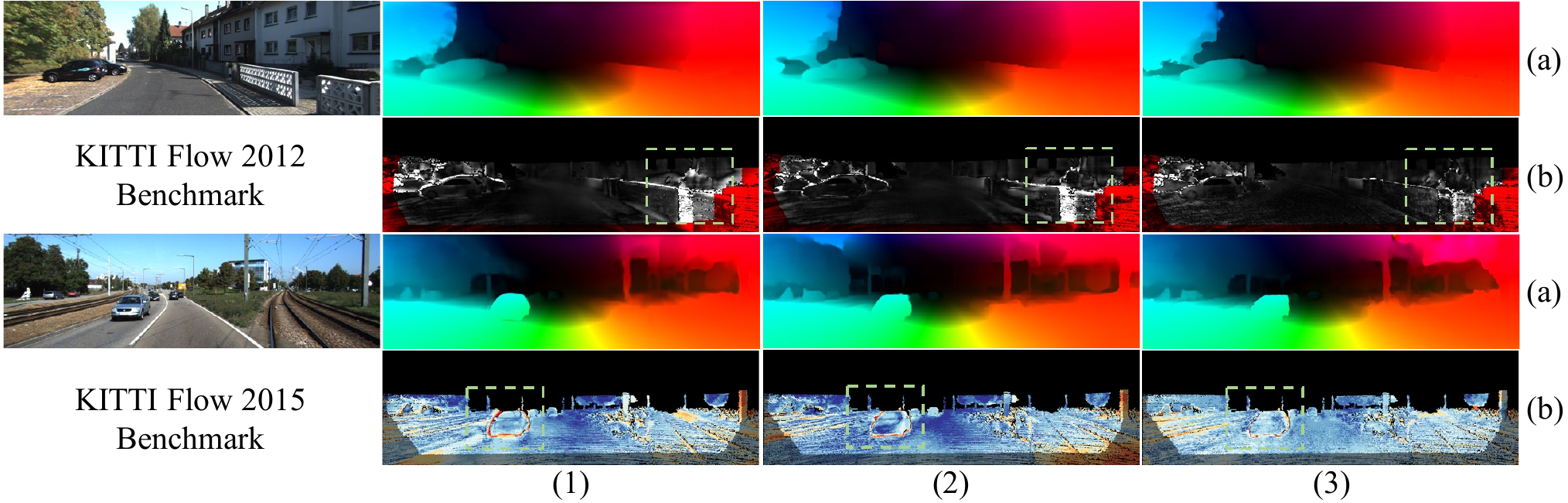}
    \caption{Examples from the KITTI flow benchmarks, where rows (a) and (b) on columns (1)--(3) show the optical flow estimations and the corresponding error maps of (1) UnFlow \cite{meister2018unflow}, (2) SelFlow \cite{liu2019selflow} and (3) our SwiftFlow, respectively. Significantly improved regions are highlighted with green dashed boxes.}
    \label{fig.kitti}
\end{figure}

\begin{figure}[!t]
    \centering
    \includegraphics[width=\textwidth]{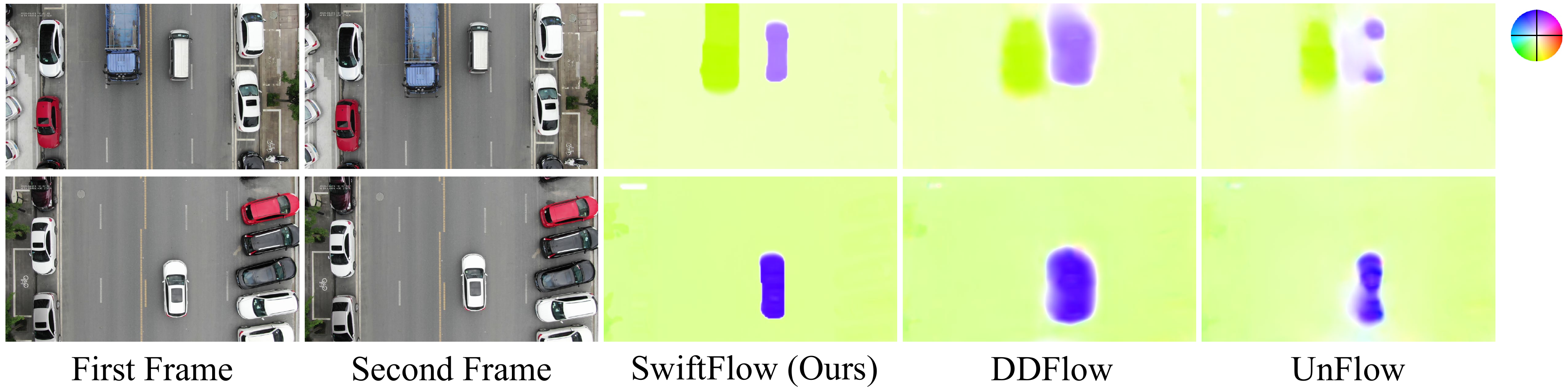}
    \caption{Examples of the optical flow estimation results on our ATG-PVD dataset. Our proposed SwiftFlow is compared with DDFlow \cite{liu2019ddflow} and UnFlow \cite{meister2018unflow}.}
    \label{fig.flow_comparison}
\end{figure}

Since our ATG-PVD dataset does not contain optical flow ground truth, we evaluate our proposed SwiftFlow on the KITTI flow 2012 \cite{kitti12} and 2015 \cite{kitti15} benchmarks. According to the online leaderboard of the KITTI flow benchmarks, as shown in Table \ref{tab.kitti}, our SwiftFlow ranks 24th on the KITTI flow 2012 benchmark\footnote{\footnotesize{\url{cvlibs.net/datasets/kitti/eval_stereo_flow.php?benchmark=flow}}} and 35th on the KITTI flow 2015 benchmark\footnote{\footnotesize{\url{cvlibs.net/datasets/kitti/eval_scene_flow.php?benchmark=flow}}}, {outperforming all other state-of-the-art unsupervised optical flow estimation approaches, with a faster running speed (in real time) achieved in the mean time}. Fig. \ref{fig.kitti} presents examples from the KITTI flow benchmarks, where we can see that SwiftFlow yields more robust results than others. Furthermore, Fig. \ref{fig.flow_comparison} shows optical flow estimation results on our ATG-PVD dataset, indicating that our proposed SwiftFlow performs much more accurately than both DDFlow \cite{liu2019ddflow} and UnFlow \cite{meister2018unflow}, another two well-known unsupervised optical flow estimation approaches, especially on the boundary of the MCs.

\subsection{Evaluation of Flow-RCNN}
\label{sec.evaluations_of_parking_violator_candidate_detection}

In our experiments, we compute the mean average precision (mAP) over ten IoU thresholds between 0.50 and 0.95 (refer to \cite{Lin2014MSCOCO} for more details) to quantitatively evaluate the performance of our proposed Flow-RCNN.  It should be noted that IPCs are regarded as IPC candidates in both training and testing experiments, due to the fact that IPCs are re-identified as IPC candidates.

We compute mAP for all three categories (IPC candidate, MC and LPC) so as to comprehensively evaluate the performance of our proposed Flow-RCNN. The quantitative results are provided in Table~\ref{tab.map_det}, where it can be observed that  Flow-RCNN outperforms the baseline network Faster-RCNN \cite{FasterRCNN} (especially for MC detection) in terms of both car detection and classification. It is rather astonishing that Faster-RCNN can still successfully detect many MCs from only RGB images, even without using optical flow information. We speculate that the baseline network might also consider the road textures around a car when inferring its category. For instance, an MC is typically at the center of a lane, and the road textures around it are similar, which can weaken the influence caused by motion blur problem.

\begin{figure}[!t]
    \centering
    \begin{tabular}{ccc}
        \includegraphics[width=.26\linewidth]{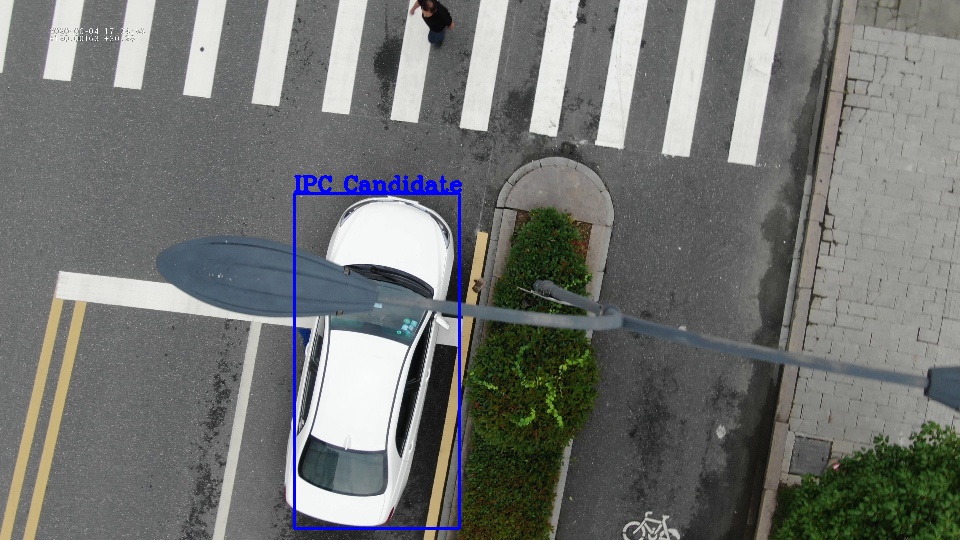} \ \ &
        \includegraphics[width=.26\linewidth]{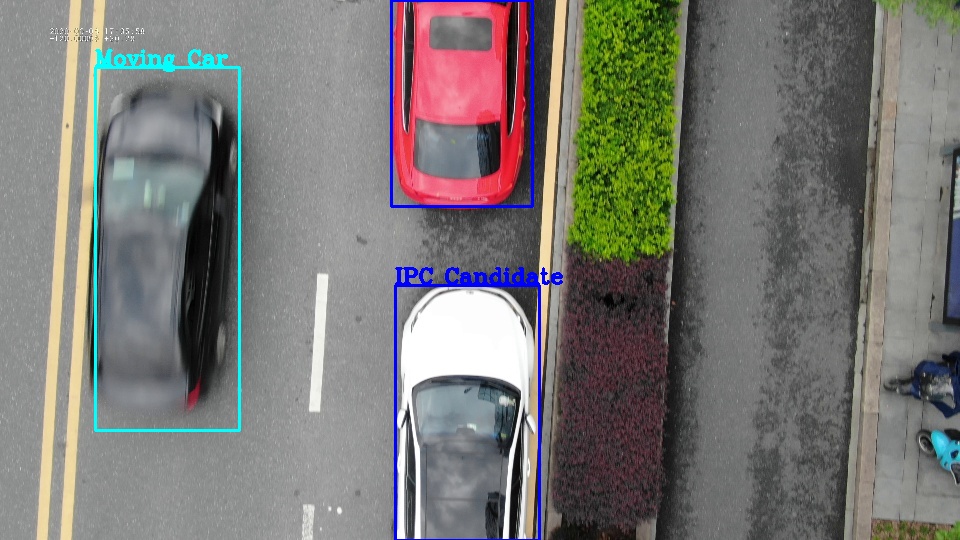} \ \ &
        \includegraphics[width=.26\linewidth]{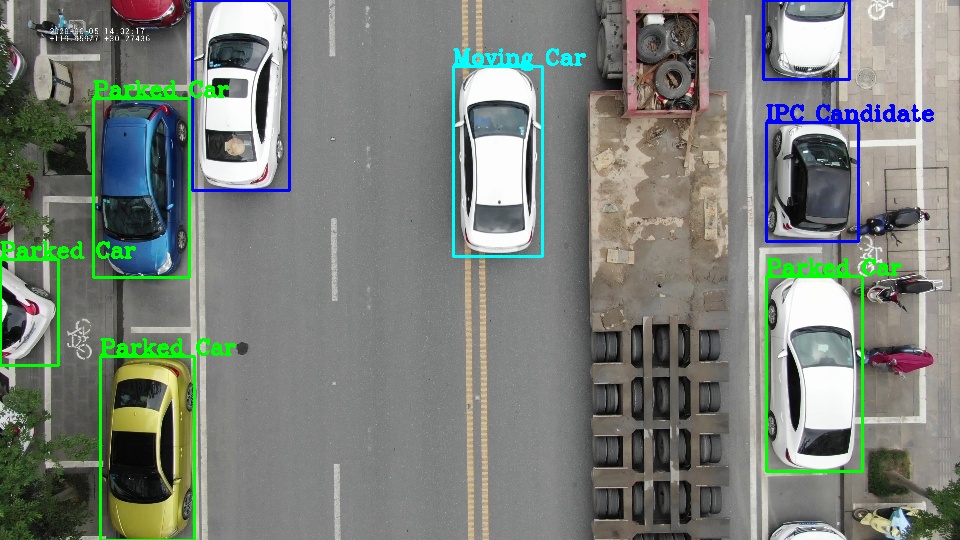} \\
        \small (a) Occlusion & \small (b) Motion blur &
        (c) Complex environment \\
    \end{tabular}
    \caption{Examples of our Flow-RCNN results.}
    \label{fig.qualitative_det}
\end{figure}

Experimental results of our Flow-RCNN are given in Fig.~\ref{fig.qualitative_det}, showing the robustness of our proposed approach. For example, in Fig.~\ref{fig.qualitative_det}(a), the light pole, that occludes part of an IPC candidate, can produce a similar optical flow estimation to an MC. Fortunately, our Flow-RCNN which fuses both RGB and flow information can still detect the IPC candidate correctly. Furthermore, although it is hard to extract features from a blurred car image, it can be seen in  Fig.~\ref{fig.qualitative_det}(b) that our proposed approach can avoid such misdetections by leveraging additional optical flow information. Moreover, in complex environments, such as the case shown in  Fig.~\ref{fig.qualitative_det}(c),  car with different categories can still be successfully detected and classified.

\begin{table}[!t]
    \caption{Car detection mAP, where the best results are shown in bold font.}
    \centering
    \begin{tabular}{L{3.1cm}C{2.1cm}C{2.1cm}C{2.1cm}C{2.1cm}}
        \toprule
        Method & Total & IPC candidate & MC & LPC \\
        \midrule
        Faster-RCNN \cite{FasterRCNN} & 0.770 & 0.844 & 0.672 & 0.789 \\
        Flow-RCNN (ours) & \textbf{0.789} & \textbf{0.845} & \textbf{0.733} & \textbf{0.796} \\
        \bottomrule
    \end{tabular}
    \label{tab.map_det}
\end{table}

\subsection{Evaluation of Parking Violation Detection}
\label{sec.evaluations_of_the_parking_violator_detection_system}
\begin{figure}[!t]
    \centering
    \includegraphics[width=0.9\textwidth]{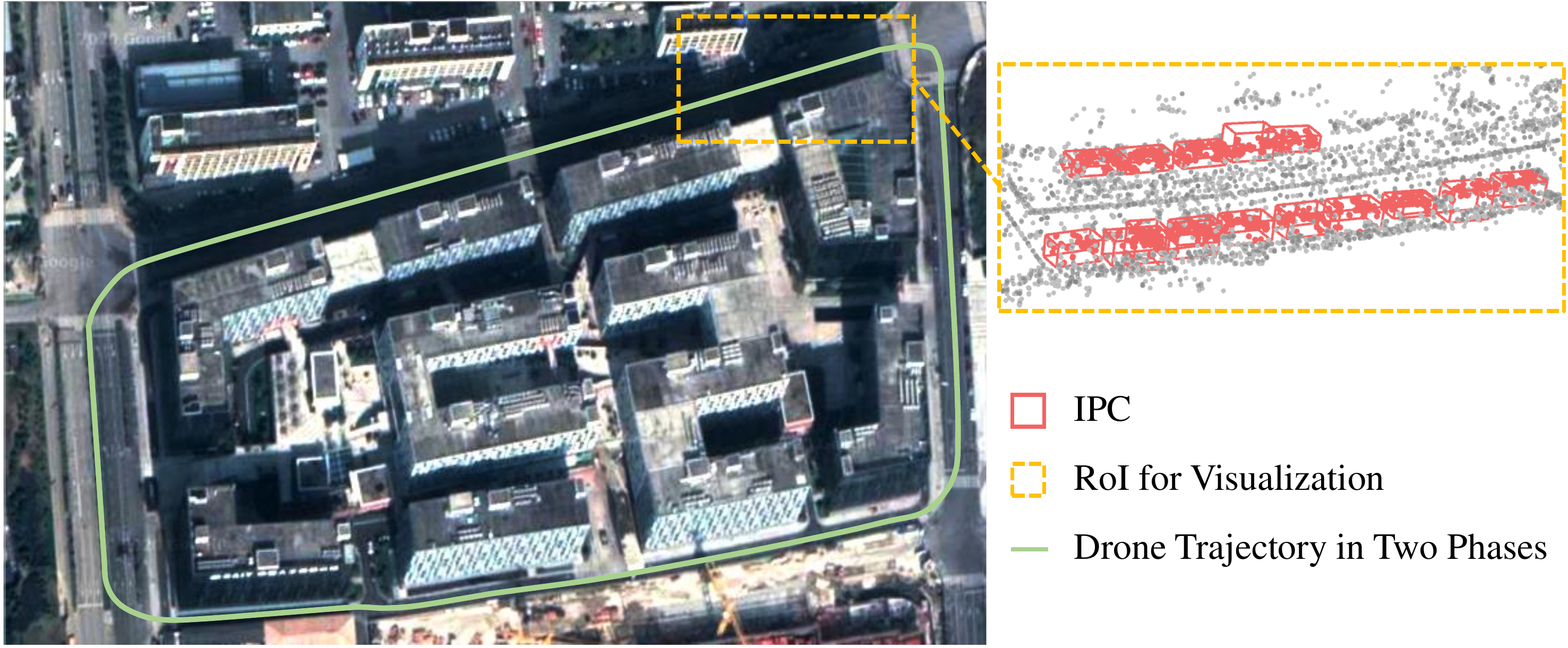}
    \caption{An example of the detected IPCs in the map.}
    \label{fig.demo}
\end{figure}

We also comprehensively evaluate the performance of the entire system for parking violation detection using our ATG-PVD dataset, and a precision of $91.7\%$, a recall of $94.9\%$ and an F1-Score of $93.3\%$ are achieved. An example of the detected IPCs in the map is illustrated in Fig.~\ref{fig.demo}, where readers can observe that our proposed suspect-and-investigate system can detect parking violations effectively and efficiently.

\section{Conclusion}
\label{sec.conclusion_and_future_work}
In this paper, we proposed a novel, robust and cost-effective parking violation detection system embedded in an ATG-R680 drone equipped with a TX2 GPU. Our system utilizes a so-called suspect-and-investigate framework, which consists of: 1) an unsupervised optical flow estimation network named SwiftFlow, 2) a novel flow-guided object detection network named Flow-RCNN, and 3) a drone re-localization and IPC re-identification module based on VSLAM. On the KITTI flow 2012 and 2015 benchmarks, our proposed SwiftFlow outperforms all other state-of-the-art unsupervised optical flow estimation approaches in terms of both speed (real-time performance was achieved) and accuracy. By incorporating the inferred optical flow information into our object detection framework, IPC candidates, MCs and LPCs can be effectively detected and classified, even in many challenging cases. In the investigation phase, our VSLAM module detects loop closure to re-localize the drone in the pre-built map.
After the drone is successfully re-localized, we further re-identify whether an existing IPC candidate is an actual IPC. The experimental results both qualitatively and quantitatively demonstrate the effectiveness and robustness of our proposed parking violation detection system.

\section*{Acknowledgements}
This work was supported by the National Natural Science Foundation of China, under grant No. U1713211, Collaborative Research Fund by the Research Grants Council Hong Kong, under Project No. C4063-18G, and the Research Grant Council of Hong Kong SAR Government, China, under Project No. 11210017, awarded to Prof. Ming Liu.

\clearpage
\bibliographystyle{splncs}
\bibliography{egbib}
\end{document}